\documentclass[10pt,twocolumn,letterpaper]{article}

\usepackage{cvpr}
\usepackage{times}
\usepackage{epsfig}
\usepackage{graphicx}
\usepackage{amsmath}
\usepackage{amssymb}
\usepackage{bbm}
\usepackage{ulem}
\usepackage{booktabs}
\usepackage[dvipsnames]{xcolor}
\usepackage[belowskip=-15pt,aboveskip=3pt]{caption}
\usepackage{enumitem}
 
\newcommand*{\mydot}{\makebox[1ex]{\textbf{$\cdot$}}}
\newcommand*{\im}[2]{\mathcal{#1}_{#2}}%
\usepackage[breaklinks=true,bookmarks=false]{hyperref}

\cvprfinalcopy 

\def\cvprPaperID{****} 

\begin{document}

\title{Improving Semantic Segmentation through Spatio-Temporal Consistency Learned from Videos}


\author{Ankita Pasad$^1$\thanks{Work done while at Robotics at Google.}\\
{\tt\small ankitap@ttic.edu}
\and
Ariel Gordon$^2$\\
{\tt\small gariel@google.com}
\and
Tsung-Yi Lin$^2$ \\
{\tt\small tsungyi@google.com}
\and
Anelia Angelova$^2$\\
{\tt\small anelia@google.com}
\\ \\
$^1$ Toyota Technological Institute at Chicago \\
$^2$ Robotics at Google }

\maketitle

\begin{abstract}

We leverage unsupervised learning of depth, egomotion, and camera intrinsics to improve the performance of single-image semantic segmentation, by enforcing 3D-geometric and temporal consistency of segmentation masks across video frames. The predicted depth, egomotion, and camera intrinsics are used to provide an additional supervision signal to the segmentation model, significantly enhancing its quality, or, alternatively, reducing the number of labels the segmentation model needs. Our experiments were performed on the ScanNet dataset.
\end{abstract}

\vspace{-2mm}
\section{Introduction}
The computer vision community has seen immense progress in solving a variety of semantic image understanding tasks, such as classification and segmentation. Typically, a deep convolutional network learns to predict labels from pixels, remaining mostly unaware of the geometric and physical constraints that govern the visual world.

Learning from video streams, as opposed to images, offers temporal coherency as a strong cue that can significantly enhance segmentation. These cues are often utilized \cite{Oh_2019_ICCV} through dedicated network architectures, capable of both segmenting and correlating objects in time.

\begin{figure}[t]
\centering
\includegraphics[width=8.5cm, trim = {0 0 0 1.8in}, clip]{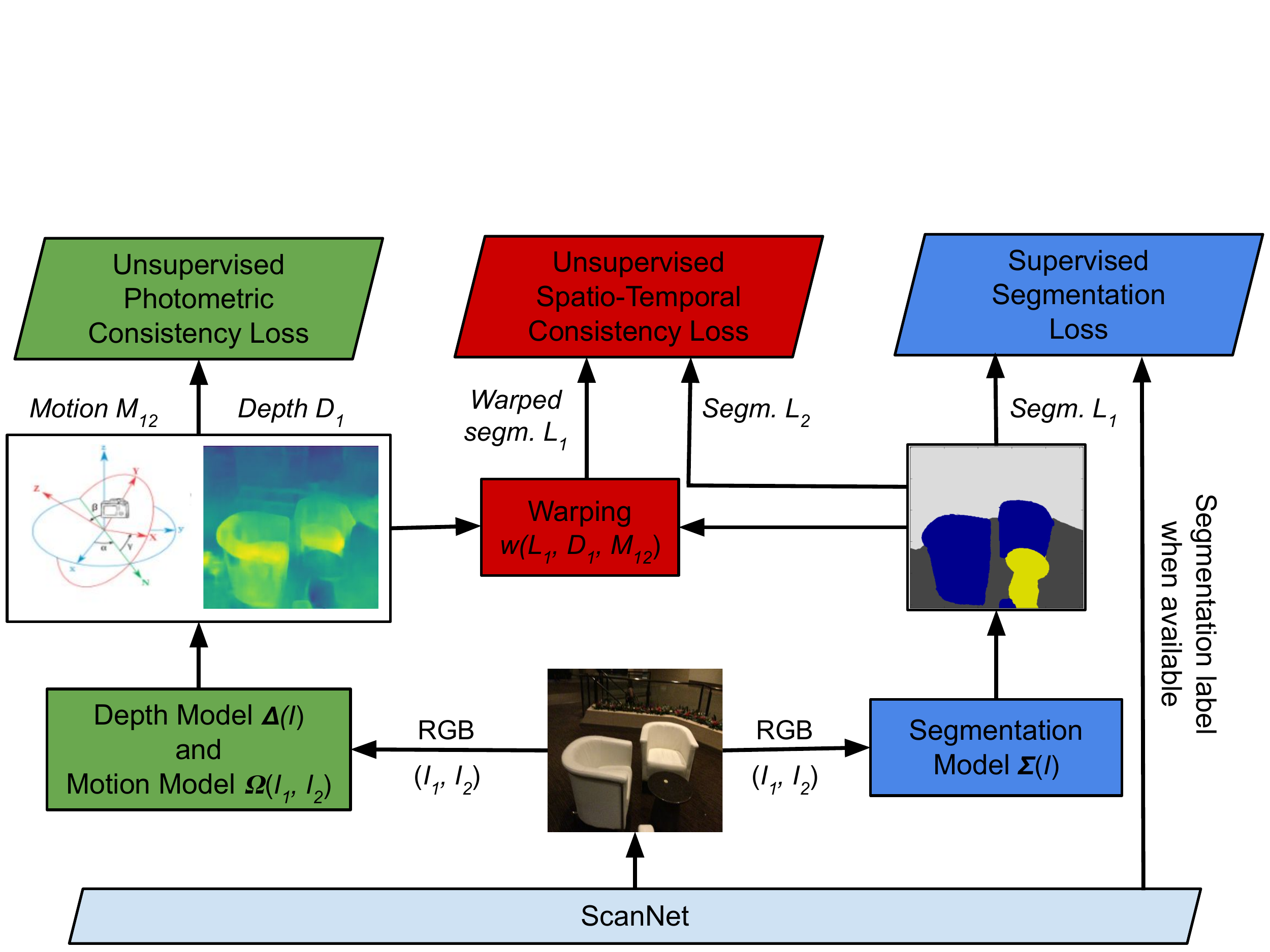}
\caption{\small \label{fig:model} Illustration of the proposed consistency-driven training of the segmentation model.}
\end{figure}

3D Multiview consistency is another cue, shown \cite{guerry2017snapnet, Ma2017MultiviewDL} to improve semantic segmentation, both as an additional supervision signal to train a single-frame segmenter, and as an additional signal at multi-frame inference time. However, these methods generally require RGBD inputs. 

Recent advances in unsupervised depth and egomotion estimation can bring together temporal continuity and multiview consistency as supervision signals for improving segmentation models. A depth prediction model can transform video sequences to RGBD sequences, correspondences between pixels in neighboring frames can be established, and consistency in segmentation mask across corresponding pixels can be used as a supervision signal. Together with the recently-demonstrated \cite{Gordon_2019_ICCV} ability to learn the camera intrinsics from unlabeled videos, depth and egomotion prediction networks facilitate adapting techniques that were previously reserved to RGBD input to general unlabeled video.

In this work, we demonstrate the effectiveness of 4D consistency constraints (that is, 3D multiview consistency alongside with temporal continuity) in providing an additional supervision signal for a semantic segmentation task. The latter is trained in a semi-supervised manner, that is, with only a small fraction of the segmentation labels used, whereas the depth, motion and intrinsics models needed for asserting the 4D consistency are trained fully unsupervised. By only using previously-published off-the-shelf networks, we demonstrate that our approach does not require any network architecture tuning, lending itself to future improvements by upgrading the respective models as they become available. The focus of our study can be summarized in three questions:
\begin{enumerate}[noitemsep,topsep=0pt,leftmargin=*]
    \item{How to construct a training objective that enforces spatio-temporal consistency?}
    \item How much improvement can the spatio-temporal (4D) consistency constraints provide when training a single-image semantic segmentation model?
    \item{How many labels can the consistency constraints replace?}
\end{enumerate}
The study is performed on the ScanNet \cite{dai2017scannet} dataset. 

\section{Related Work}
The advantages of learning geometric and semantic tasks jointly have been long recognized \cite{eigen2015predicting, standley2019tasks, zou2018df}. Eigen and Fergus \cite{eigen2015predicting} are among the first to exploit relations between geometry and semantics by learning them jointly. Their work provides empirical evidence for performance improvement when optimizing depth and semantics jointly. Right-left geometric consistency has been used to improve semantic segmentation in a stereo setting \cite{chen2019towards, jiang2019sense}. Optical flow has been used to improve the temporal consistency of segmentation masks \cite{hur2016joint,nilsson2018semantic}. To the best of our knowledge, this work is the first attempt at improving single-image segmentation by employing spatio-temporal consistency using unsupervised depth, egomotion, and camera intrinsics estimates learned from videos.

\section{Method}\label{sec:consistency_losses}
In this section we discuss the formulation of consistency losses that are based on the idea of self-supervision and the design does not assume access to any task specific supervision. We start with a pair of consecutive images from a video, $\im{I}{t}$ and $\im{I}{t+1}$. We have access to the models for depth estimation, camera motion estimation, and semantic segmentation: $\Delta(\mydot)$, $\Omega(\mydot, \mydot)$, and $\Sigma(\mydot)$ respectively. We first estimate the depth $d_t=\Delta(\im{I}{t})$ and the camera motion, i.e., 3D rotation and translation, $M_{t,t+1} = \Omega(\im{I}{t}, \im{I}{t+1})$. Simultaneously, the segmentation model generates the logits masks, $L_t = \Sigma(\im{I}{t})$ and $L_{t+1} = \Sigma(\im{I}{t+1})$. Using the depth and motion estimates we have a differentiable warping function, $\omega(\mydot, \mydot, \mydot)$ that gives us an estimated transformation function from $\im{I}{t}$ to $\im{I}{t+1}$. We thus have an additional estimate for the logits mask $\hat{L}_{t+1} = \omega(L_t, d_t, M_{t, t+1})$. An overview of the proposed method is presented in fig.~\ref{fig:model}. 

Now, we employ consistency between the propagated logits mask, $\hat{L}_{t+1}$, and the predicted logits mask, $L_{t+1}$. Note that the proposed loss formulation will hold for the backward consistency constraint between $\hat{L}_{t}$ and $L_t$ as well, where $\hat{L}_{t} = \omega(l_{t+1}, d_{t+1}, M_{t+1, t})$.
\begin{align}
\label{eq:l1_loss}
\ell_{L1} = \sum_{x,c}W(x, c)||\hat{L}_{t+1}(x, c) - L_{t+1}(x, c)||_1
\end{align}
where $x$ is the pixel index in 2D space, $c$ is the class index, $W(\mydot, \mydot)$ is the normalized weight for the L1 difference as a function of pixel location and class label. We use a combination of 3 different formulations of the weighing function with the respective weights as 0.2, 0.4, and 0.4.
\begin{enumerate}[noitemsep,topsep=0pt,leftmargin=*]
    \item {Uniform}: Mean of the difference, where $W(x, c)$ is constant across both pixel and class indices in the mask.
    \item Label prior: Uniform averaging fails to differentiate between the classes that actually appear in the image from those that don't, whereas it is more reasonable to have a higher penalty for the inconsistencies in the former. Since we do not have the groundtruth labels we use $\hat{y}_{t+1}$ as a belief for the same and set $W(x, \mydot) = \mathbbm{1}_{\hat{y}_{t+1}(x)}$, where $\mathbbm{1}_{\hat{y}_{t+1}(x)}$ is a one-hot vector of the length same as the number of classes with 1 at index $\hat{y}_{t+1}(x)$, and $\hat{y}_{t+1} = \text{argmax} (L_{t+1})$ along the class index axis.
    \item Pixel prior: Here, the weight is constant across different classes while the inconsistencies for edge pixels are penalized more than the others. Here, $W(\mydot, c) = E(\im{I}{t+1}) \text{ }\forall\text{ }c$, where $E(\im{I}{t+1})$ is the two-dimensional edge detector output for the image, with one for pixel locations corresponding to the edges. 
\end{enumerate}

\section{Experiments}
\subsection{Models} All models trained in our experiments -- depth prediction, egomotion prediction, and semantic segmentation -- were taken from other publications, using their respective open-sourced code, along with their tuned optimization hyperparameter settings. This choice allows gauging the quality improvements associated with imposing consistency constraints, as opposed to architectural improvements. By applying the consistency constraints in an architecture-agnostic manner, we leave an open route to further improvements, by simply replacing the comprising models by better ones, as they become available in the literature. For semantic segmentation we use the NAS-FPN \cite{ghiasi2019fpn} as the backbone architecture with the segmentation classifier design as proposed by Kirillov et al. \cite{kirillov2019panoptic}. For the prediction of depth, egomotion, and camera intrinsics we use the models from Ref.~\cite{Gordon_2019_ICCV}. Both the off-the-shelf models we use are recently proposed, strong models attaining the state of the art performance in the respective tasks.

\subsection{Dataset}
We use ScanNet \cite{dai2017scannet}, a dataset of indoor RGB-D video sequences. It consists of 2.5M views across 1500 scans. All the frames in a video sequence are labeled for semantic segmentation masks across 21 classes including a background class. The annotations were obtained by rendering the 3D scans from the sequence of 2D images to get 1500 3D scans. These 3D scans were then manually annotated for segmentation and were projected back to 2D. This is a good dataset for the proof of concept evaluation of our method as it gives a handle to freely control the available supervision to create an artificial limited supervision setting. 

\subsection{Key Results}
\begin{table}
\small
\begin{center}
\begin{tabular}{|l||c|c|c|c|c|c|}
\hline
\% supervision & 0.1 & 0.2 & 0.5 & 1 & 2 & 4 \\
\hline
Baseline & 39.8 & 43.2 & 47.0 & 49.0 & 49.1 & 51.1 \\ \hline
With consistency & 43.3 & 46.1 & 49.4 & 49.3 & 51.1 & 51.1 \\ \hline
\end{tabular}
\end{center}
\caption{\small \label{table:results} Mean intersection over union (MIOU) scores (\%) for semantic segmentation on ScanNet validation set, for models trained with varied fraction of the labels (\% supervision), with and without spatio-temporal consistency. No improvement in the MIOU was observed above 4\% supervision due to the strong correlations between the ScanNet images.}
\vspace{2mm}
\end{table}
The effect of spatio-temporal consistency on the segmentation performance is summarized in Table \ref{table:results}. The fraction of labels used for training is varied from 0.1\% to 4\%, where the rest of the images were stripped of their labels and only used for imposing spatio-temporal consistency. At or above 4\% supervision, we observe a mean intersection over union (MIOU) of 51.1\%\footnote{This number is in the ballpark of prior RGB-only segmentation benchmarks of MIOU=50.3\% \cite{valada2019self}, however the numbers are not directly comparable since our results were evaluated on the validation set rather than the test set, and because different subsets of the training set were used.}. For each case, the supervised baseline model is obtained by training the single-image segmentation model on the respective labeled data. The MIOU of the resulting baseline model is summarized in the second row of Table~\ref{table:results}. Depth, egomotion, and camera instrinsics models were trained separately, unsupervised, on the entire ScanNet training dataset. Then the consistency loss in Eq.~\ref{eq:l1_loss} was switched on as an additional supervision signal for the segmentation model, and the latter continued to train, achieving the MIOU values summarized in the third row of Table~\ref{table:results}. We notice that the proposed approach gives consistent improvements across the range of supervision. Higher relative improvements are observed at lower supervision. 

Additionally, it is meaningful to analyze the effect of consistency as an alternative to direct supervision. Indeed, supervision obtained through consistency constraints can mimic an increase in the number of labels by up to a factor of \textit{four}: The baseline MIOU at 2\% supervision (49.1\%) matches the MIOU with \textit{only quarter} as much labeled data (49.4\%) with the proposed approach. 

\subsection{Effect on the Rare Labels}
\begin{figure}[t]
\centering
\includegraphics[width=8cm, trim = {0.1in 0in 0.08in 0in}, clip]{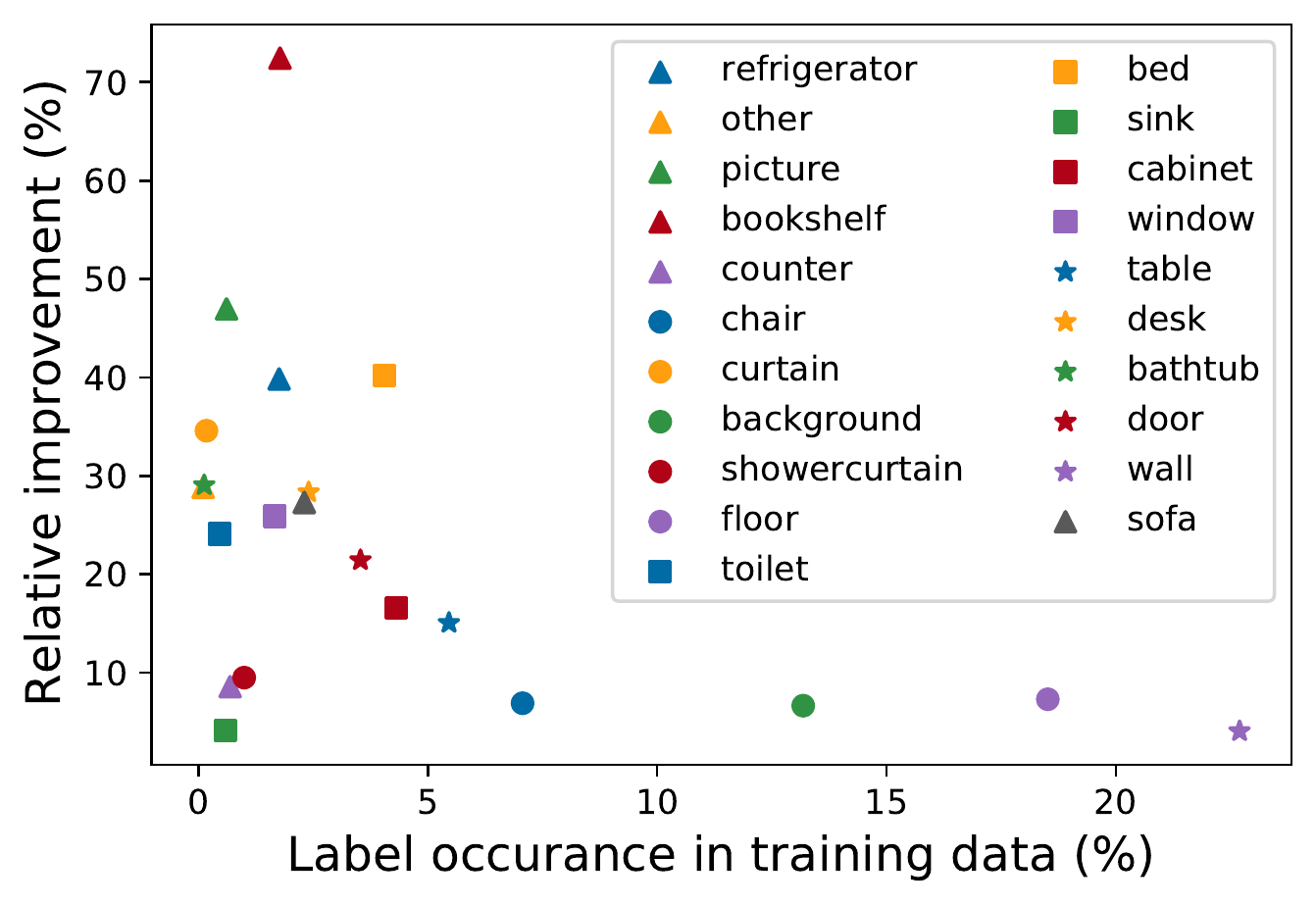}
\caption{\small Relative improvement in the class-wise MIOU score as a function of the class frequency in the labeled data.}
\label{fig:freq_plot}
\centering
\vspace{3mm}
\includegraphics[width=8.5cm, trim = {0.15in 2.2in 2.5in 0}, clip]{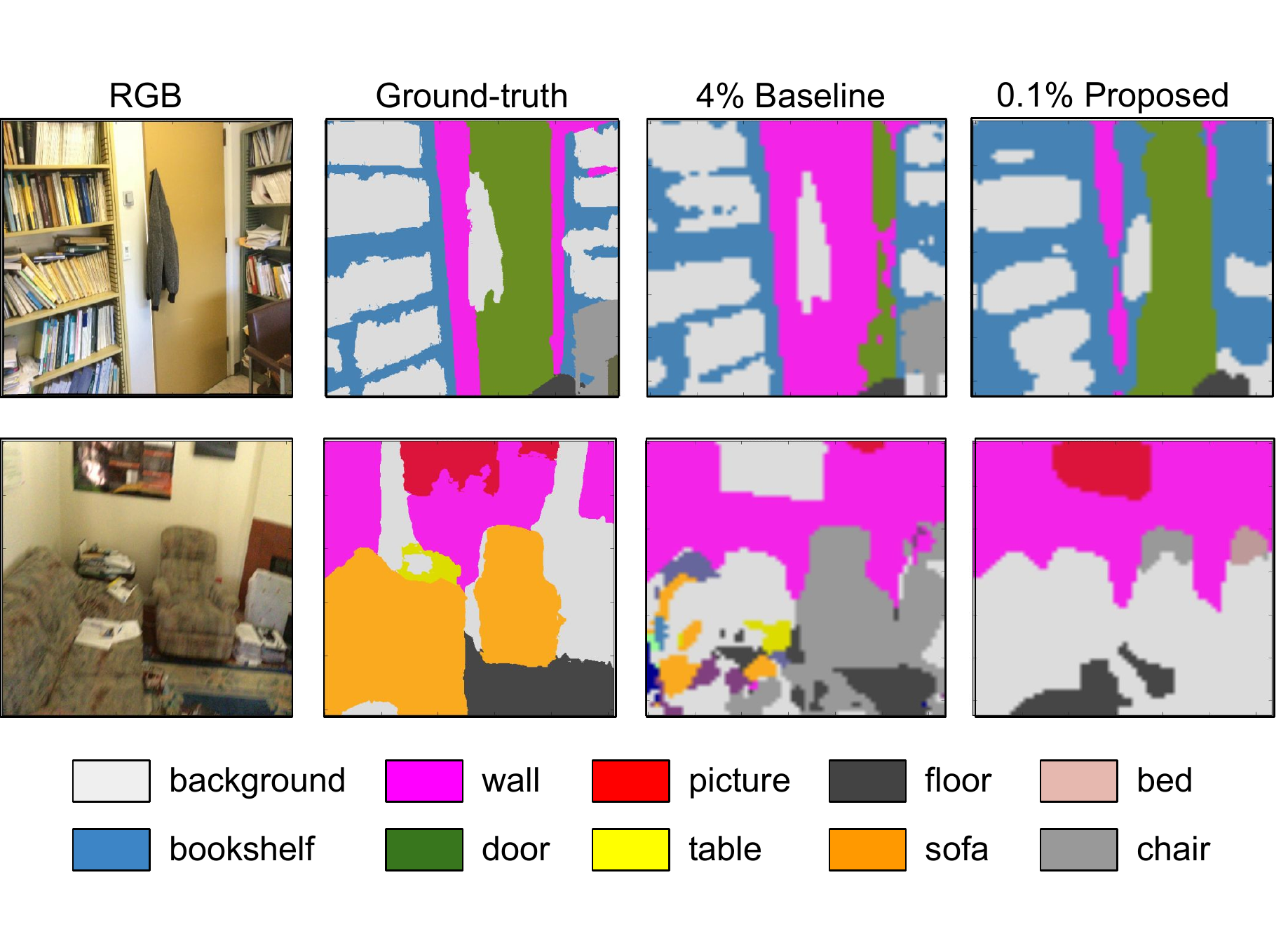}
\caption{\small \label{fig:qualitative}Qualitative comparison of the 4\% baseline with the proposed approach trained on 0.1\% supervision. Rare labels ``door" (top) and ``picture" (bottom) are successfully identified by the proposed approach.}
\vspace{3mm}
\end{figure}

Fig.~\ref{fig:freq_plot} presents the relative improvement in per-class MIOU for the 0.1\% baseline when trained with the proposed consistency constraints. We observe that the relative improvement in the MIOU tends to be the highest for the rarest labels, especially those that are only 5\% frequent or lesser. The spatio-temporal consistency constraints effectively augment the training labels to include more instances of each label. This naturally leads to improved performance for especially less frequent labels.

Fig.~\ref{fig:qualitative} demonstrates two such examples where the best baseline fails to identify masks for rare labels, ``door" and ``picture", whereas the proposed consistency-based approach trained with \textit{40 times less direct supervision} is able to accurately label the corresponding pixels. The examples also show that the predictions for the proposed model are smoother across pixels corresponding to a particular object with hardly any fragmentation artifacts.

\subsection{Ablation Study of consistency constraints}
Table~\ref{table:ablation} presents the individual contribution of different components of the consistency loss across rows 2 to 5. We can see that each loss components helps individually while the combination of the three performs the best. Additionally the last two rows of the table are added for reference where the logits are used to generate the pseudo labels. These pseudo labels are then used to train the warped logits using the cross-entropy loss $\ell_{CE} = CE(\hat{L}_{t+1}, \hat{y}_{t+1})$. We see that this straight-forward approach based on the hard decision on predictions is not nearly as good as the proposed averaging method to impose the consistency. 
\begin{table}
\small
\begin{center}
\begin{tabular}{|c|c|c||c||c|}
\hline
Uniform & Label Prior & Pixel Prior & $\ell_{CE}$& MIOU \\
\hline\hline
\multicolumn{4}{|c|}{baseline} & 39.8 \\ \hline
\checkmark &  &  &  & 41.7 \\ \hline
 & \checkmark &  &  & 41.3 \\ \hline
& & \checkmark  &  & 41.8 \\ \hline
\checkmark & \checkmark & \checkmark &  & {\bf 43.3} \\ \hline \hline
 &  &  & \checkmark & 41.4 \\ \hline
\checkmark & \checkmark & \checkmark & \checkmark & 42.0 \\ \hline
\end{tabular}
\end{center}
\caption{\small \label{table:ablation}Effect of different consistency losses (detailed in Sec.~\ref{sec:consistency_losses}) on the segmentation performance.}
\end{table}

In all the experiments above, the depth, egomotion, and camera intrinsics models supervised the segmentation model via spatio-temporal consistency losses, but not the other way around. Allowing the segmentation model to supervise the other models resulted in no significant improvement in the depth estimation error. While such improvements were observed in prior work \cite{Gordon_2019_ICCV}, they were attributed to the ability of segmentation to identify moving objects. This ability is irrelevant to ScanNet's static scenes. 
The analyses and the ablation studies presented in this section were done on the 0.1\% supervision case.

\section{Conclusion}
In this work, we used models predicting depth, egomotion, and camera intrinsics, to provide additional supervision to a semantic segmentation model through spatio-temporal consistency constraints. The latter were shown to reduce the need for direct supervision by a factor of up to four. Enhancement in semantic segmentation performance was observed, especially for the less frequent labels. All models were adopted from prior publications, through our approach that is network-architecture-agnostic.

The method proposed in this work can be readily extended to dynamic scenes. Rather than only estimating camera motion, dynamic scenes require the estimation of 3D object motion relative to the scene. It has been previously shown \cite{Gordon_2019_ICCV} that segmentation can provide a regularization for 3D motion estimation. The consistency losses developed in this study can provide supervision from the depth and motion model to the segmentation model, closing the loop on the three models, depth, 3D motion and segmentation, peer-supervising each other. 

{\small
\bibliographystyle{ieee_fullname}
\bibliography{bib_file}
}

\end{document}